\documentclass[lettersize,journal]{IEEEtran}
\usepackage{amsmath,amsfonts}
\usepackage{array}
\usepackage[caption=false,font=normalsize,labelfont=sf,textfont=sf]{subfig}
\usepackage{textcomp}
\usepackage{stfloats}
\usepackage{url}
\usepackage{verbatim}
\usepackage{graphicx}
\usepackage{cite}
\usepackage{adjustbox}
\usepackage{xcolor}
\usepackage{pifont}
\usepackage{amssymb}
\usepackage{multirow}
\newcommand{\cmark}{\ding{51}} 

\usepackage[ruled, lined, linesnumbered, commentsnumbered, longend]{algorithm2e}
\usepackage[colorlinks, linkcolor=green,
anchorcolor=green,
citecolor=green]{hyperref}
\begin{document}
\title{Edge-guided Representation Learning for Underwater Object Detection}

\author{Linhui Dai, Hong Liu, Pinhao Song, Hao Tang, Runwei Ding, Shengquan Li
\thanks{Linhui Dai and Hong Liu are with the Key Laboratory of Machine Perception, Shenzhen Graduate School, Peking University, Shenzhen, China (Email: dailinhui@pku.edu.cn, hongliu@pku.edu.cn).}
\thanks{Pinhao Song is with the Robotics Research Group, KU Leuven, Leuven, Belgium (Email: pinhao.song@kuleuven.be).}  
\thanks{Hao Tang is with the Computer Vision Lab, ETH Zurich, Zurich, Switzerland (Email: hao.tang@vision.ee.ethz.ch).} 
\thanks{Runwei Ding is with the Peng Cheng Laboratory, Shenzhen, China (Email: dingrw@pcl.ac.cn).} 
\thanks{Shengquan Li is with the Peng Cheng Laboratory, Shenzhen, China (Email: lishq@pcl.ac.cn).} 
}

\maketitle
\begin{abstract}
	Underwater object detection (UOD) is crucial for marine economic development, environmental protection, and the planet's sustainable development. The main challenges of this task arise from low-contrast, small objects, and mimicry of aquatic organisms. The key to addressing these challenges is to focus the model on obtaining more discriminative information. We observe that the edges of underwater objects are highly unique and can be distinguished from low-contrast or mimicry environments based on their edges. Motivated by this observation, we propose an Edge-guided Representation Learning Network, termed ERL-Net, that aims to achieve discriminative representation learning and aggregation under the guidance of edge cues. Firstly, we introduce an edge-guided attention module to model the explicit boundary information, which generates more discriminative features. Secondly, a feature aggregation module is proposed to aggregate the multi-scale discriminative features by regrouping them into three levels, effectively aggregating global and local information for locating and recognizing underwater objects. Finally, we propose a wide and asymmetric receptive field block to enable features to have a wider receptive field, allowing the model to focus on more small object information. Comprehensive experiments on three challenging underwater datasets show that our method achieves superior performance on the UOD task.
\end{abstract}
\section{Introduction}

Nowadays, underwater object detection (UOD) has attracted great attention since its significance in marine economic development and environmental protection \cite{zhao2021composited}. UOD is a vital computer vision task aiming locate and recognize marine organisms or marine debris. And it can be applied in many areas, such as scientific breeding and fishing, seafood aquaculture, and marine waste recycling. 

\begin{figure}[t]
	\centering
	\includegraphics[width=0.48\textwidth]{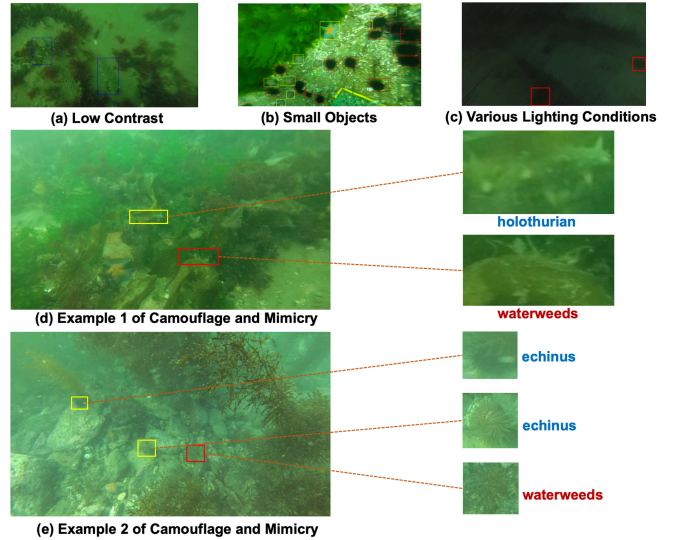}
	\caption{Some challenging examples in underwater object detection. The examples illustrate several characteristics of underwater datasets. The proposed method is designed to address these challenges.}
	\label{fig:keysight}
\end{figure}

Recently, thanks to advances in deep learning-based methods, generic object detection (GOD) has achieved tremendous improvement. The mainstream object detection methods can be categorized into two categories: the one-stage object detection methods \cite{yolov3,SSD,FCOS,retinanet, dai2022ao2} and the two-stage object detection methods \cite{FasterRCNN, FastRCNN, LibraRCNN, CascadeRCNN}. As we all know, the two-stage detectors maintain a higher accuracy of localization and classification, whereas the one-stage detectors achieve higher detection speed. However, directly applying these methods to UOD has very limited performance due to several challenges: 

(1) The contrast of underwater images is shallow, and the image is affected by the light and may have different colors at different times, as shown in Figure \ref{fig:keysight}(a) and (c); (2) Underwater objects vary in size and often include many small objects, as shown in Figure \ref{fig:keysight}(b); (3) To obtain the advantages and avoid the disadvantages is the instinct of species, undoubtedly, aquatic organisms always want to share a similar color or pattern to their preferred environment and try to blend into its environment to make them more difficult to spot. For instance, echinus and holothurian always like to live near waterweeds or holes. And example of these challenges is shown in Figure \ref{fig:keysight}(d) and (e). 

There have been many underwater object detection works that have achieved good performance. For instance, Song et al. \cite{song2023boosting} proposed a two-stage underwater detector Boosting R-CNN to improve the performance of UOD from two perspectives: uncertainty modeling and hard example mining. Fan et al. \cite{FER} proposed an underwater detection framework with feature enhancement and anchor refinement to solve sample imbalance. These existing methods \cite{song2023boosting, EfficientDets, chen2023achieving} have promoted the progress of underwater object detection, but there are still limitations. For example, they have not fully considered the shape variability of underwater objects, which limits the performance of underwater object detection in cases where small and camouflaged underwater organisms exist, as well as in low-contrast underwater environments.

Different from existing UOD methods, we address the above challenges lies in the guidance of edge cues, the fusion of global and local information by regrouping the features, and allowing features to have a wider receptive field. We propose a novel framework called Edge-guided Representation Learning network (ERL-Net) for underwater object detection. To solve the challenge (1) and (2), it is necessary to obtain a more discriminative feature representation under the guidance of edge cues. We first introduce an Edge-Guided Attention (EGA) module to model explicit boundary information. And then, a Feature Aggregation (FA) module is proposed to aggregate hierarchical features by regrouping the extracted features into low-level, mid-level, and high-level. Through the edge-guided global and local information, we can more accurately locate and identify underwater objects in low-contrast and mimicry environments. To address the challenge (3), motivated by the human visual recognition system, we present the Wide and Asymmetric Receptive Field Block (WA-RFB) to allow the features to have a wider receptive field, which is beneficial for locating small objects. With these three dedicated components, the proposed ERL-Net can mitigate complicated underwater challenges and achieve promising performance on the UOD task.

In conclusion, the main contributions of this paper can be described as follows:
\begin{itemize}
	\item We propose a novel framework called Edge-guided Representation Learning Network (ERL-Net) for underwater object detection. ERL-Net comprises three components: an edge-guided attention module, a feature aggregation module, and a wide and asymmetric receptive field block. 
	\item The core sight of the proposed ERL-Net is to model the edge information explicitly. Under the guidance of edge cues, we propose a Feature Aggregation (FA) module to aggregate global and local information, and a Wide and Asymmetric Receptive Field Block (WA-RFB) is proposed to allow the features to have a wider receptive field and excavate more discriminative contextual information. 
	\item Extensive experiments on three challenging underwater datasets, including UTDAC2020 \cite{chen2022swipenet}, Trashcan \cite{hong2020trashcan}, and Brackish \cite{brackish}, demonstrate that the proposed ERL-Net outperforms most other object detection methods and achieves state-of-the-art performance in underwater object detection.
\end{itemize}

\begin{figure*}[t]
	\centering
	\includegraphics[width=\textwidth,height=\textheight,keepaspectratio]{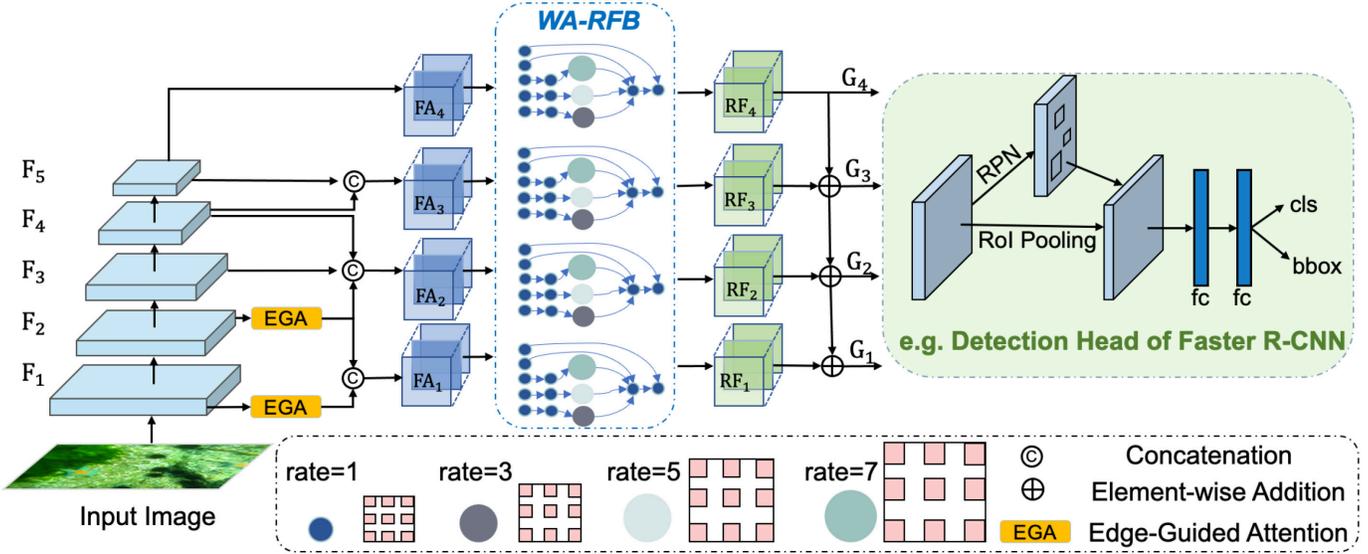} 
	\caption{Illustration of our proposed framework. Our framework is composed of the Feature Aggregation module (FA), a Wide and Asymmetric Receptive Field Block (WA-RFB), an Edge-Guided Attention mechanism (EGA), and a detection head at the end. FA is used to aggregate multi-level information. The WA-RFB is introduced to simulate the structure of RFs in the human visual system and guide the framework to pay more attention to contextual formation. The EGA is proposed to utilize edge information to guide the object features to perform better on localization. At last, a detection head can help to achieve detection in an end-to-end manner.}
	\label{fig:framework}
\end{figure*}


\section{Related Works}
In this section, we discuss three types of works related to UOD: generic object detection, salient object detection, and underwater object detection.
\subsection{Generic Object Detection}
The current generic object detection (GOD) is mainly separated into two mainstreams: one-stage methods and two-stage methods. The one-stage detectors mostly followed the works based on YOLO \cite{yolov3} or SSD \cite{SSD}. RetinaNet \cite{retinanet} proposed the focal loss to apply a modulating term to the cross-entropy loss for dealing with class imbalance. Tian et al. \cite{FCOS} proposed an anchor-free and proposal-free one-stage method FCOS to solve the object detection in a per-pixel prediction fashion. FCOS utilized multi-level prediction and the ``centerness'' branch to improve the performance. On the other hand, the two-stage methods \cite{RCNN,FastRCNN,CascadeRCNN,cascaderpn,FasterRCNN,D2Det,pafpn,ga} have a proposal-driven mechanism. In this way, these methods have more precision prediction than one-stage methods. A series of detection methods,  R-CNN \cite{RCNN}, Fast R-CNN \cite{FastRCNN}, and Faster R-CNN \cite{FasterRCNN}, rely on the region-based network. The detection consists of two stages: the model first proposes a set of regions of interest (ROI), then classifies and regresses them. Feature pyramid network (FPN) \cite{FPN} was proposed to exploit multi-scale features by incorporating a top-down path to sequentially combine features at different scales, significantly improving the performance of detecting small objects. Qiao et al. \cite{detectors} proposed a recursive feature pyramid (RFP) to incorporate extra feedback connections from FPN into bottom-up backbone layers. However, state-of-the-art object detection methods are unsuitable for UOD since the aquatic organisms are often blurred with the background and often hide around waterweeds or rocks. Thus addressing UOD requires more visual perception and attention knowledge.
\subsection{Salient Object Detection}
Salient Object Detection (SOD) aims at highlighting and segmenting visually salient object regions in an image. Previous methods \cite{des, iso} for SOD mainly rely on low-level hand-crafted features, such as appearance similarity, background prior, and region contrast. Recently, deep convolutional neural networks have attracted wide attention. For example, Zhai et al. \cite{BBSNet} proposed BBS-Net for the RGB-D SOD by leveraging the characteristics of multi-level cross-modal features and further introduced a depth-enhanced module in BBS-Net to enhance the depth features. Fan et al. \cite{COD} proposed a simple and effective framework SINet which contains a search and identification module for camouflaged object detection. Besides, EGNet \cite{EGNet} was proposed to handle the problem of coarse object boundaries in the SOD by modeling the salient edge information. Despite SOD and UOD being different in terms of task and content, SOD can provide constructive ideas for UOD. Motivated by BBS-Net, edge guidance is introduced in our method. Hence, the deep model can capture finer object features and obtain more accurate localization.

\subsection{Underwater Object Detection}
Recently, UOD has garnered much research attention considering its wide applications in marine engineering and aquatic robotics. From the perspective of data augmentation, Lin et al. \cite{ROIMIX} proposed an augmentation method called ROIMIX to conduct proposal-level fusion among multiple images for UOD. To improve the small object detection accuracy, Chen et al. \cite{chen2022swipenet} proposed SWIPENET which consists of high-resolution and semantic rich hyper feature maps, and also proposed a sample-weighted re-weighting algorithm. Fan et al. \cite{FER} proposed an underwater detection framework with feature enhancement and anchor refinement. To handle the domain shift in the UOD task, DG-YOLO \cite{liu2020towards} firstly proposed the task of underwater object detection for domain generalization. Chen et al. \cite{chen2023achieving} proposed a training paradigm named DMCL to sample new domains on the domain manifold by interpolating paired images on the feature level. To solve the low-contrast of the underwater environment, Yeh et al. \cite{yeh2021lightweight} proposed to jointly train the color conversion and object detection to improve the performance of underwater object detection.

However, the performance of these methods is still limited due to insufficient basic features extracted. Unlike the above research, we aim to design a more powerful network for extracting feature representation and exploiting the edge information to guide prediction.


\section{The Proposed Method}
\subsection{Overview}
The overall pipeline of ERL-Net is shown in Figure \ref{fig:framework}. For an input image $\mathbf{I} \in \mathbb{R}^{H \times W \times 3}$, a set of features $\left\{F_k\right\}_{k=0}^5$ is extracted from ResNet-50 \cite{resnet50}. The resolution of each layer is $\left\{\left[\frac{H}{k}, \frac{W}{k}\right], k=4,8,16,32,64\right\}$. Our goal is to model explicit boundary information. Then, under the guidance of edge cues, we aggregate the hierarchical global and local features by regrouping the features into three levels. Next, we use the WA-RFB to allow the features to have a wider receptive field. Finally, a detection head is used to output the classification and localization results. All components will be detailed in the following sections.

\subsection{Edge-guided Attention Module}
For UOD, underwater objects have diverse appearances, which is the embodiment of biodiversity. Therefore, the key to extracting contextual information is to use these distinctly different edges. Unlike other edge attention modules \cite{BBSNet, EGNet} applied to semantic segmentation, the proposed edge-guided attention (EGA) module is the first to directly excavate edge information by using deep Sobel convolution in UOD.

\begin{figure}[t]
	\centering
	\includegraphics[width=0.48\textwidth,keepaspectratio]{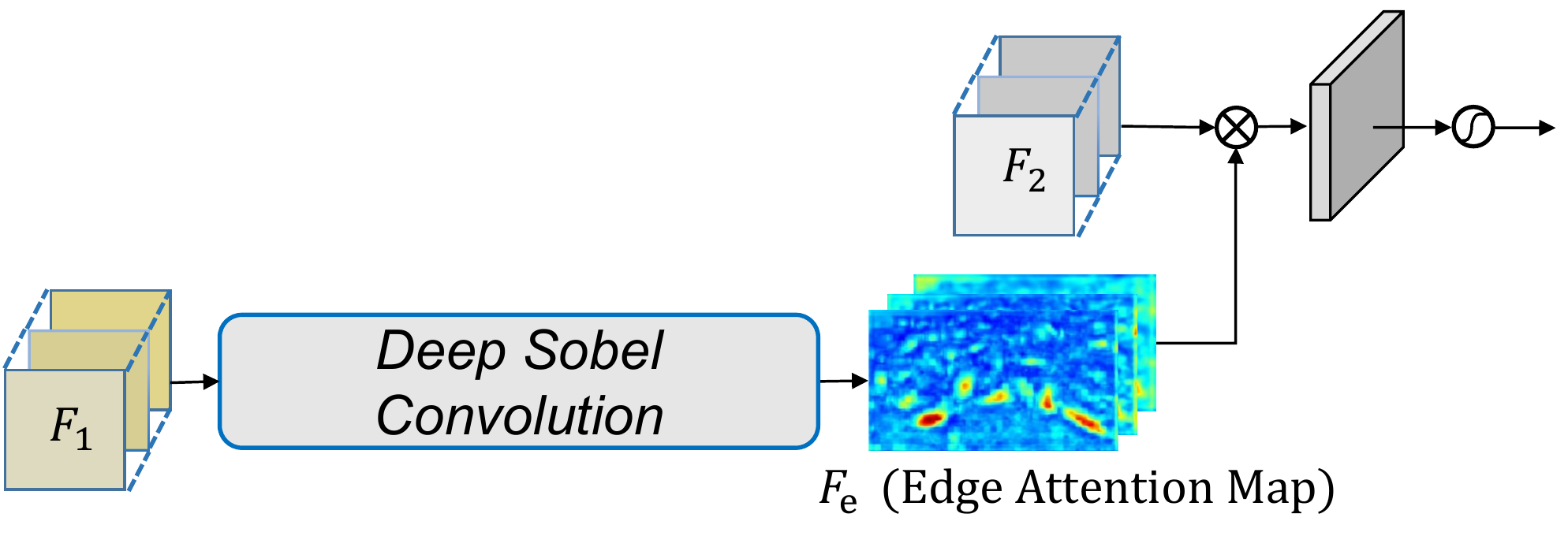} 
	\caption{The pipeline of edge-guided attention module (EGA). It takes the feature map from Conv1 as input. The edge feature map is obtained by deep Sobel convolution and then multiplied by $C_2$. EGA is utilized to learn edge features implicitly and then fuse the complementary edge information and object information.}
	\label{fig:ega_pipeline}
\end{figure}

\begin{figure}[t]
	\centering
	\includegraphics[width=0.48\textwidth, keepaspectratio]{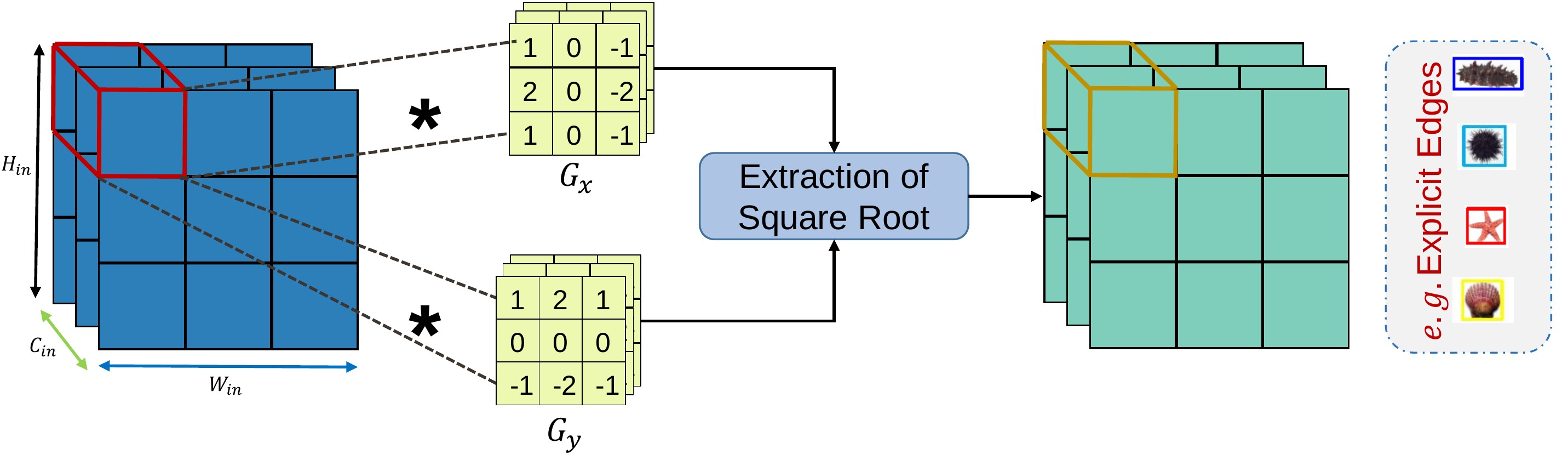}
	\caption{The illustration of deep Sobel convolution is a process of modeling the explicit edges of underwater objects. The EGA takes the feature map from $C_1$ as input, multiplied by the two $3 \times 3$ Sobel kernels $(G_x, G_y)$, respectively.}
	\label{fig:EGA}
\end{figure}

Considering that the low-level features preserve sufficient edge information, the feature $F_1$ of Conv1 is fed to the proposed edge attention module to produce the edge attention map. Figure \ref{fig:ega_pipeline} shows a detailed architecture of the EGA. In details, we obtain the edge attention map $\widetilde{F}_{1}$ and $\widetilde{F}_{2}$ by multiplying (element-wise, $\odot$) the low-level features $F_1$ and saliency representation of edge attention map $F_e$. To compute $F_e$, we design a deep Sobel kernel $(G_x, G_y)$ to apply on every channel of shadow feature maps, as illustrated in Figure \ref{fig:EGA}. The formula of edge features is defined as:
\begin{equation}
\widetilde{\mathbf{F}}_{1}=F_{1} \odot F_{e1}, \\
\widetilde{\mathbf{F}}_{2}=F_{2} \odot F_{e2}, \\
\end{equation}

\begin{equation}
	G_{x}=\left[\begin{array}{ccc}
	1 & 0 & -1 \\
	2 & 0 & -2 \\
	1 & 0 & -1
	\end{array}\right], G_{y}=\left[\begin{array}{ccc}
	1 & 2 & 1 \\
	0 & 0 & 0 \\
	-1 & -2 & -1
	\end{array}\right],
\end{equation}
\begin{equation}
	F_{ei}=\sqrt{\left(\left(\operatorname{grad}_{xi}\right)^{2}+\left(\operatorname{grad}_{yi}\right)^{2}\right)}, i \in \{1,2\},
\end{equation}
\begin{equation}
	\begin{aligned}
	\operatorname{grad}_{xi}&=\sum_{C_{i n}}^{c} \sum_{W_{i n}}^{w} \sum_{H_{i n}}^{h} C_{ic}(w, h)  \cdot G_x, \\
	\operatorname{grad}_{yi}&=\sum_{C_{i n}}^{c} \sum_{W_{i n}}^{w} \sum_{H_{i n}}^{h} C_{ic}(w, h)  \cdot G_y, \\
	\end{aligned}
\end{equation}
where $G_x$ and $G_y$ are two Sobel kernels, one for horizontal changes and one for vertical changes. Here, $i \in \{1,2\}$, $F_i$ denote the low-level features.  Moreover, $grad_x$ and $grad_y$ are edge response maps which at each point contain the horizontal and vertical derivative approximations, respectively. Then we take the average response map of all filters to get the total edge feature map $F_{ei}$. Finally, we can get the edge-guided features $\widetilde{F}_{1}$ and $\widetilde{F}_{2}$. 

\begin{figure}[t]  
\centering  
\includegraphics[width=0.48\textwidth,keepaspectratio]{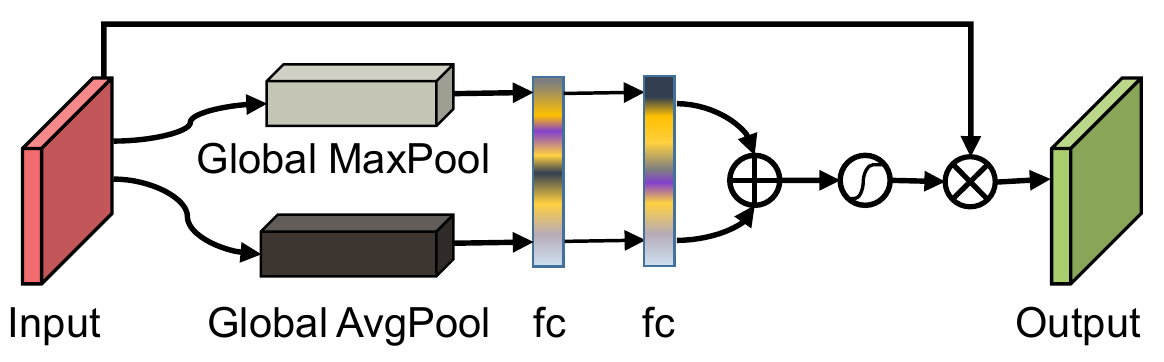}    
\caption{The architecture of channel-wise attention module. Specifically, the channel-wise attention module utilizes both the outputs of max-pooling and average-pooling, then passes through two fully-connected layers and finally combines on learned weights.} 
 \label{fig:ca} 
\end{figure}

On the other hand, the attenuation of light depends on the wavelength of light and the dissolution of organic compounds, causing the underwater images to always look bluish or greenish. It is one of the main differences between GOD and UOD. Consequently, in low-contrast underwater environments, we need to increase discriminative representation to focus on important features and suppress unnecessary ones. Evolved from \cite{cbam, SENet}, we focus on the channel relationship to adaptively recalibrate channel-wise feature responses by explicitly modeling interdependencies between channels, and then ``what'' is meaningful can be noticed. Figure \ref{fig:ca} shows the architecture of channel-wise attention.
 
For aggregating and inferring fine spatial information, we combine both global average pooling and max pooling to generate channel-wise statistics. The channel-wise attention is a computational unit which can be built upon a transformation $F_{tr}$ mapping an input $\mathbf{F} \in \mathbb{R}^{H^{\prime} \times W^{\prime} \times C^{\prime}}$ to feature maps $\mathbf{U} \in \mathbb{R}^{H \times W \times C}$. Formally, a statistic $\mathrm{z} \in \mathbb{R}^{C}$ is generated by shrinking $U$ through its spatial dimensions $H \times W$. We first aggregate the spatial information of the feature maps by using average pooling and max pooling operations, generating two different spatial context descriptors: $F_{\text{avg}}^{\mathrm{c}}$ and $F_{\text{max}}^{\mathrm{c}}$, which represent the average-pooled feature and the max-pooled feature, respectively. Then, these two descriptors are forwarded to a shared network to generate our new channel attention map $F \in \mathbb{R}^{C \times 1 \times 1}$, which is calculated by: 
\begin{equation} 
	 O=\sigma\left(W_{1} \delta\left(W_{0} F_{\text{avg}}^{\mathrm{c}} \right) \oplus W_{1} \delta\left(W_{0} F_{\text{max}}^{\mathrm{c}}\right)\right) \otimes X, 
\end{equation} 
where $\sigma$ denotes the sigmoid function, $\delta$ refers to the ReLU function, $W_0 \in \mathbb{R}^{C / r \times C}$ and $W_1 \in \mathbb{R}^{C \times \frac{2}{r}}$ are shared for both inputs and the ReLU activation function. Here, $\oplus$ denotes the element-wise addition, and $\otimes$ denotes the element-wise multiplication.

\subsection{Multi-level Feature Aggregation}
According to the knowledge of evolutionary biology, underwater organisms always camouflage themselves in the environment, which is a way of achieving crypsis. It allows otherwise visible aquatic organisms to remain unnoticed by other organisms such as predators or prey. For instance, in Figure \ref{fig:keysight} (d) and (e), holothurian and echinus are most likely to hide in similar waterweeds. If we only focus on the local information of the objects, we are likely to get wrong prediction results. It has shown that multi-level information is important for locating and classifying objects \cite{COD, FPN}. The high-level features in the deep layers contain rich semantic information for locating objects, while the low-level features in the shallow layers preserve spatial details for determining the boundary of objects. For an input image $\mathbf{I} \in \mathbb{R}^{W \times H \times 3}$, the features $\left\{F_{k}\right\}_{k=1}^{5}$ are extracted from ResNet-50 \cite{resnet50}. To model the explicit edge information, we use the edge-guided low-level features $\widetilde{\mathbf{F}}_{1}$ and $\widetilde{\mathbf{F}}_{2}$ to replace the original low-level features. Based on the above considerations, we divide the extracted features into low-level \{$\widetilde{F}_{1}$, $\widetilde{F}_{2}$\}, middle-level \{$F_3$\}, and high-level \{$F_4$, $F_5$\}, then combine them through concatenation, up-sampling, and down-sampling operations. The resolution and channel size of features $\{\widetilde{F}_1, \widetilde{F}_2, F_3, F_4, F_5\}$ are  $\left\{\left[\frac{H}{k}, \frac{W}{k}\right], k=4,8,16,32,64\right\}$ and $\{c_1, c_2, c_3, c_4, c_5\}$, respectively. As shown in Figure \ref{fig:framework}, after aggregating the above multi-level features, we have a set of hierarchical aggregated features \{$FA_1$, $FA_2$, $FA_3$, $FA_4$\}, the corresponding channel size set is $\{c_1 + c_2, c_2 + c_3 + c_4, c_4 + c_5, c_5\}$. Hierarchical global and local information can be beneficial for locating and recognizing underwater objects.

\subsection{Wide and Asymmetric Receptive Field Block} 
\begin{figure}[t]
	\centering
	\subfloat[RFB]{
	  \label{fig:subfig:a}
	  \includegraphics[scale=0.35]{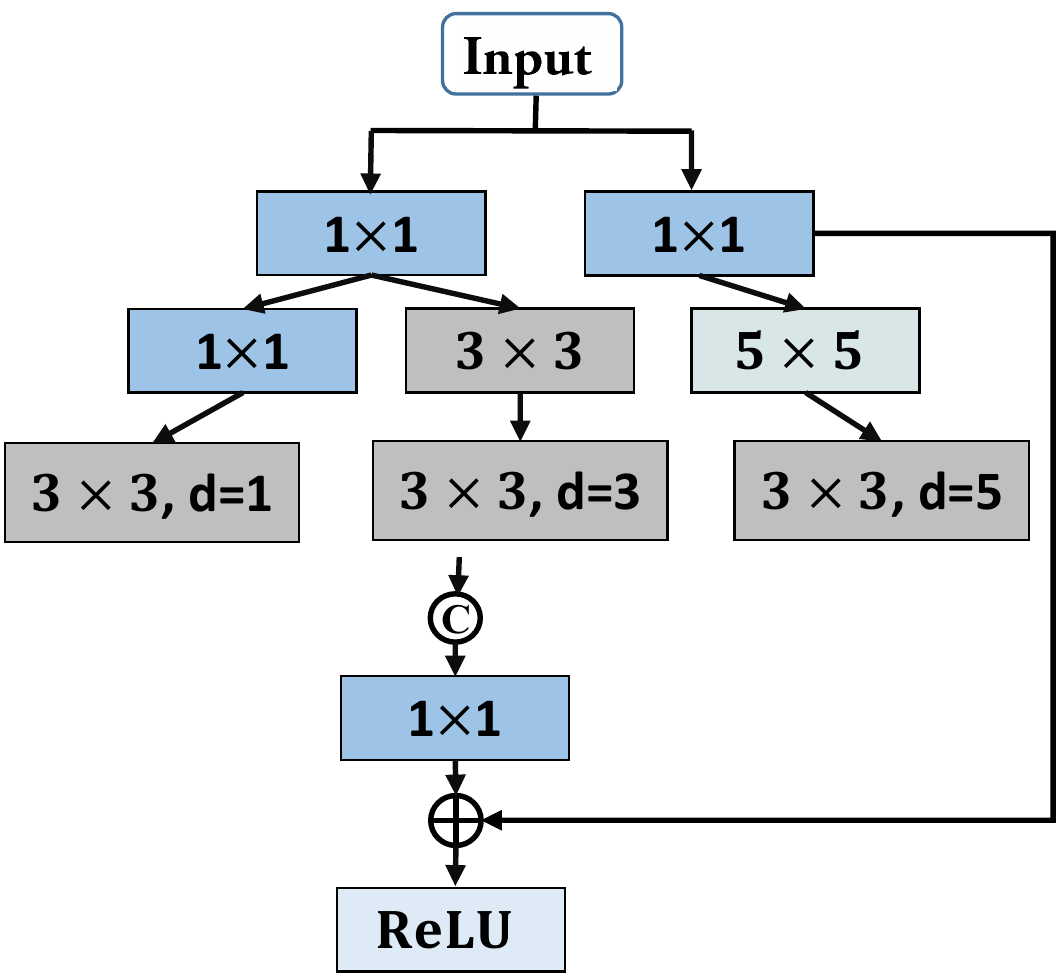}}
	\subfloat[WA-RFB]{
	  \label{fig:subfig:b} 
	  \includegraphics[scale=0.35]{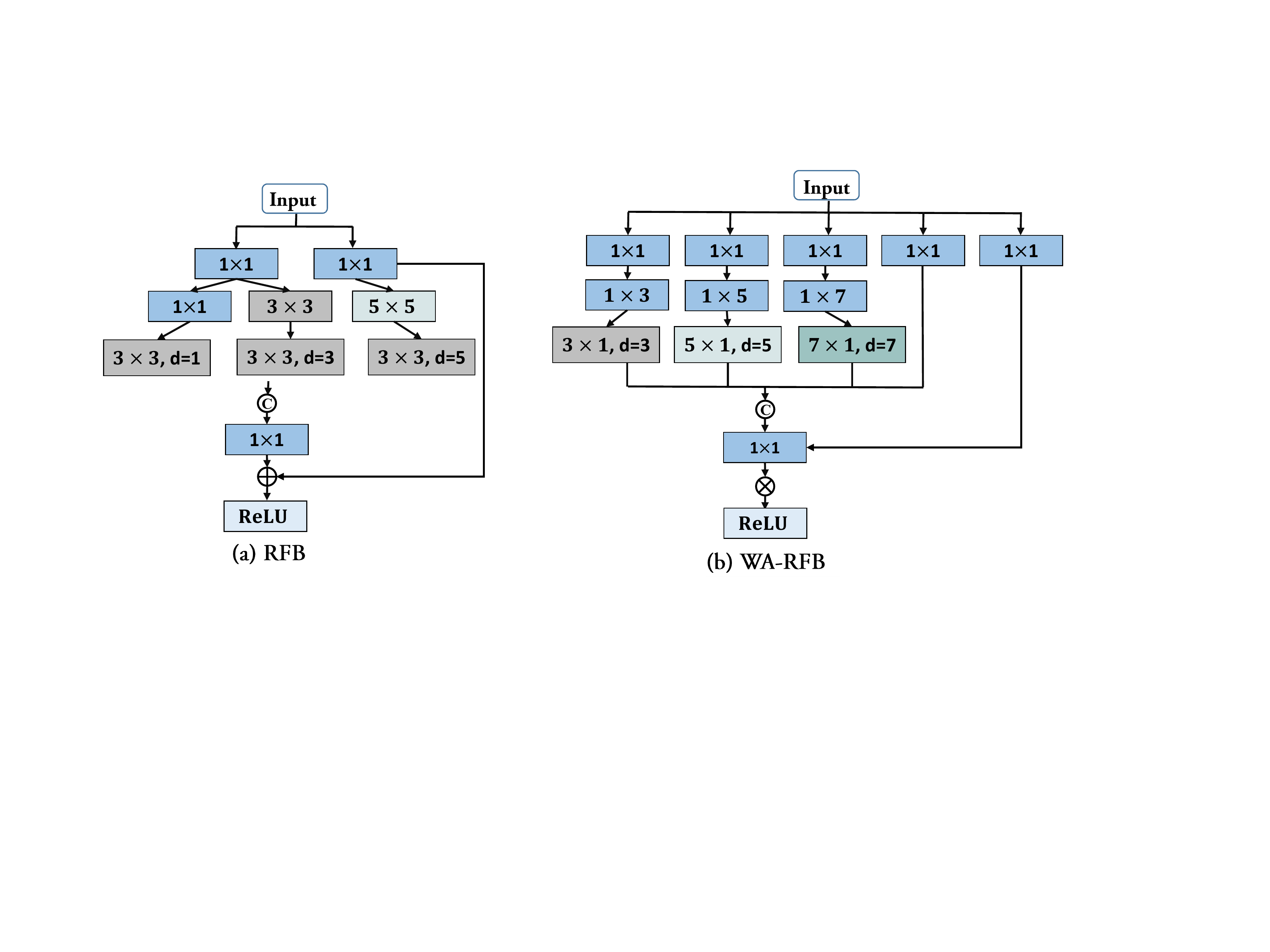}}
	\caption{The architecture of the proposed Wide and Asymmetric Receptive Field Block (WA-RFB) and the foundation work of RFB \cite{RFB}. WA-RFB consists of 5 branches with different small kernels. It can enrich multi-scale discriminative contextual information which is critical for underwater object detection.}
	\label{fig:irf} 
\end{figure}

In GOD, most of the objects occupy a large area and are easy to identify. However, in UOD, the objects only occupy a small part of the images, and they are too dense and small to be recognized. The key to detecting small objects is to allow features to have a wider receptive field and excavate more discriminative contextual information. To effectively excavate the contextual information, we use the relevant knowledge of the human visual perception system to incorporate more discriminative feature representation. Based on the work of RFB \cite{RFB}, we propose a Wide and Asymmetric Receptive Field Block (WA-RFB). Figure \ref{fig:irf} shows the architecture of WA-RFB. Different from RFB, WA-RFB contains five branches  $\left\{b_{k}, k=1, \ldots, 5\right\}$ and has different ways of feature concatenation. In particular, WA-RFB is an asymmetric structure that focuses on discriminative information of different importance. In each branch, the first convolutional layer has a dimension of $ 1 \times 1$ to reduce the channel size to 32. This is followed by two other layers: a $ 1 \times (2k-1)$ convolutional layer and a $(2k-1) \times 1$ convolutional layer with a specific dilation rate $(2k-1)$ when $(1<k<5)$. The first four branches are concatenated, and then followed with a $1 \times 1$ convolutional operation, respectively. Finally, the $5^{th}$ branch is multiplied with the concatenated features, and the whole module is fed to a ReLU function to obtain the feature $RF_{k}$.

WA-RFB highlights the relationship between the size and eccentricity of receptive fields, and forces the network to learn discriminative information through an asymmetric structure, which is beneficial to recognize small samples. The whole process can be expressed as:
\begin{equation}
	X_{out}=\tau\left(Br_5 \otimes \epsilon(Br_1 \odot Br_2 \odot Br_3 \odot Br_4)\right),
\end{equation}
where $X_{out}$ represents the output feature, $Br_1$, $Br_2$, $Br_3$, $Br_4$, and $Br_5$ denote the output of the five branches, $\odot$ represents the operation of feature concatenation, and $\otimes$ represents the operation of feature multiplication. Here, $\epsilon$ denotes the process of adjusting the number of channels through $1 \times 1$ convolution, and $\tau$ is the activation function of ReLU.

Thus, the outputs of the WA-RFB module can be denoted as $\{RF_{1}, RF_{2}, RF_{3}, RF_{4}\}$. Then, we adopt the FPN \cite{FPN} architecture to output a set of feature maps $\left\{G_{i} \mid i=1, \ldots, 4\right\}$, where $i$ denote the $i^{th}$ stage of resnet-50. The output feature $G_i$ is defined by:
\begin{equation}
G_{i}=T_{i}\left(G_{i+1}, RF_{i}\right), 
\end{equation}
where $RF_{4} = {FA}_{4}$. Let $T_i$ denote the $i$-th top-down FPN operation. The $G_{i}$ is used as detection features of the detection head.

\subsection{Loss Function} 
ERL-Net can be applied to many existing popular two-stage or single-stage detectors, by simply adding the block into the backbone. After getting the final feature map sets, they will be sent to a detection head to do the localization and classification. Figure \ref{fig:framework} shows a detection head of Faster R-CNN \cite{FasterRCNN}, it first utilizes a region proposal network (RPN) in the first stage. Then RoI pooling and local regression module are used to separate classification and regression branches in the second stage. In this paper, we  conduct experiments by applying ERL-Net with RetinaNet \cite{retinanet}, Faster R-CNN \cite{FasterRCNN}, and Cascade R-CNN \cite{CascadeRCNN}.

The proposed framework is optimized in an end-to-end manner. For the two-stage pipeline, the RPN loss $\mathcal{L}_{rpn}$ and classification loss $\mathcal{L}_{cls}$ remain the same as Faster R-CNN. For the single-stage pipeline, the loss function is the same as RetinaNet.

\section{Experiments and Results}
\label{sec:experiments}
In this section, we first introduce the underwater datasets used in the experiments. Then, we provide the implementation details and evaluation metrics. Next, we present quantitative results with SOTA methods. Subsequently, we perform a detailed ablation analysis to investigate the basic modules of our method. Finally, we present the qualitative results and error analysis of our method.

\begin{table*}
	\centering
	\caption{Comparisons with state-of-the-art on the UTDAC2020 dataset. When using a ResNet50 backbone with FPN, ERL-Net achieves the best performance on the all metrics, with an overall AP of 0.484, surpassing all existing two-stage and single-stage object detectors.   The results with red and blue colors indicate the best and second-best results of each column, respectively.}
	\label{tab1}
	\begin{adjustbox}{max width=\textwidth}
	\begin{tabular}{l|ccc|ccc|cccc} 
	\hline
	Methods & AP  & $AP_{50}$ & $AP_{75}$ & $AP_S$ & $AP_M$ & $AP_L$& holothurian & echinus& scallop & starfish \\ 
	\hline
	 \textbf{\textit{Two-Stage Detector:}}     &  &&  &   & & &&& & \\
	Faster R-CNN \cite{FasterRCNN}& 0.448 & 0.810& 0.444           & 0.182           & 0.389           & 0.514           & 0.375           & 0.448           & 0.462           & 0.505            \\
	Faster R-CNN w/ FPN \cite{FPN}              & 0.469           & 0.820           & 0.491           & 0.236           & 0.416           & 0.529           & 0.385           & 0.485           & 0.481           & 0.526            \\
	Cascade RPN \cite{cascaderpn}              & 0.471           & 0.805           & 0.506           & 0.207           & 0.409           & 0.537           & 0.384           &  \textcolor{blue}{0.490}           & 0.475           & 0.534            \\
	PAFPN \cite{pafpn}                                    & 0.470           & 0.820           & 0.499           & 0.234           & 0.415           & 0.530           & 0.397           & 0.483           & 0.477           & 0.523            \\
	Cascade R-CNN w/ FPN \cite{CascadeRCNN}                     & 0.476           & 0.823           & 0.504           & 0.206           & 0.420           & 0.540           & 0.398           & 0.489           & 0.484           & 0.534            \\
	Libra R-CNN \cite{LibraRCNN}                           & 0.469           & 0.815           & 0.490           & 0.231           & 0.412           & 0.530           & 0.387           & 0.487           & 0.473           & 0.528            \\
	DetectoRS \cite{detectors}                             & 0.465           & 0.817           & 0.490           & 0.226           & 0.410           & 0.525           & 0.381           & 0.484           & 0.473           & 0.520            \\
	D2Det \cite{D2Det}                                   & 0.435           & 0.802           & 0.426           & 0.183           & 0.378           & 0.495           & 0.347           & 0.457           & 0.439           & 0.495            \\
	SABL w/ Faster R-CNN \cite{sabl}                     & 0.478           & 0.815           & 0.504           & 0.211           & 0.421           & 0.541           & 0.401           & 0.484           & \textcolor{red}{0.490}           & 0.535            \\ 
	SABL w/ Cascade R-CNN \cite{sabl}                  & 0.478           & 0.814           & 0.507           & 0.214           & 0.415           & \textcolor{red}{0.544}           & 0.402           & 0.488           & 0.485           & 0.539            \\ 
	\hline
	 \textbf{\textit{Single-Stage Detector:}}  &                 &                 &                 &                 &                 &                 &                 &                 &                 &                  \\
	RetinaNet w/ FPN \cite{retinanet}                               & 0.414           & 0.766           & 0.396           & 0.169           & 0.348           & 0.480           & 0.327           & 0.446           & 0.387           & 0.495            \\
	SABL w/ RetinaNet \cite{sabl}                               & 0.414           & 0.777           & 0.398           & 0.189           & 0.352           & 0.480           & 0.329           & 0.438           & 0.394           & 0.494            \\
	SSD \cite{SSD}                               & 0.400           & 0.775           & 0.365           & 0.147           & 0.361           & 0.451           & 0.308           & 0.419           & 0.412           & 0.462            \\
	FSAF \cite{fsaf}                                      & 0.397           & 0.758           & 0.379           & 0.168           & 0.336           & 0.464           & 0.318           & 0.402           & 0.389           & 0.479            \\
	NAS-FCOS \cite{nasfcos}                                   & 0.458           & \textcolor{blue}{0.828}           & 0.461           & 0.217           & 0.401           & 0.524           & 0.370           & 0.460           & 0.475           & 0.525            \\ 
	\hline 	 
	\textbf{\textit{Transformer:}}  && & & && &&& & \\
	Deformable DETR \cite{deformable} & 0.461   & 0.841  & 0.456 & 0.250& 0.422 & 0.513 & 0.388 & 0.470 & 0.466&0.522 \\
	Swin Transformer \cite{liu2021swin} & 0.453   & 0.828  & 0.451 & 0.221& 0.401 & 0.512 & 0.372 & 0.457 & 0.462&0.521 \\
	\hline
	 \textbf{\textit{Ours}}  & & & && &&&&& \\
	ERL w/ RetinaNet    & 0.451           & 0.808           & 0.463           & 0.205           & 0.396           & 0.506           & 0.360           & 0.484           & 0.441           & 0.519            \\
	ERL w/ Faster R-CNN  & \textcolor{blue}{0.483}           & \textcolor{red}{0.832}  & \textcolor{blue}{0.513}           & \textcolor{blue}{0.251}           & \textcolor{red}{0.430}  & \textcolor{blue}{0.542}           & \textcolor{blue}{0.405}           & \textcolor{red}{0.498}           & \textcolor{blue}{0.488}           & \textcolor{blue}{0.541}            \\
	ERL w/ Cascade R-CNN & \textcolor{red}{0.484}  & \textcolor{blue}{0.828}           & \textcolor{red}{0.522}  & \textcolor{red}{0.261}  & \textcolor{blue}{0.427}           & \textcolor{red}{0.544}  & \textcolor{red}{0.406}  & \textcolor{red}{0.498}  & \textcolor{blue}{0.488}  & \textcolor{red}{0.543}   \\
	\hline
\end{tabular}
\end{adjustbox}
\end{table*}

\begin{table} 	
\centering 	
\caption{Benchmarking results between ERL-Net and other state-of-the-art methods on Brackish dataset. The results with red and blue colors indicate the best and second-best results of each column, respectively.} 	
\label{tab2} 	
\begin{adjustbox}{max width=\textwidth} 	
\begin{tabular}{ll}  		\hline 		
Methods            & $AP_{50}$    \\  		\hline 		
YOLOv3 \cite{brackish}            & 0.837            \\ 		
Faster R-CNN w/ FPN \cite{FasterRCNN}   & 0.629          \\ 		
RetinaNet w/ FPN \cite{retinanet}            & 0.937            \\ 	
SABL w/ Faster R-CNN \cite{sabl}  & 0.986             \\ 		
SABL w/ Cascade R-CNN \cite{sabl}  & 0.987            \\ 		
D2Det \cite{D2Det}               & 0.981         \\
Boosting R-CNN \cite{song2023boosting}  &   0.974  \\
Deformable DETR \cite{deformable}    & 0.823   \\ 
Swin Transformer \cite{liu2021swin}      & 0.852      \\ 	
\hline 		
Ours  &   \\ 		
ERL w/ RetinaNet & \textcolor{blue}{0.979}  \\ 		
ERL w/ Faster R-CNN           & \textcolor{red}{0.988}  \\ 		
ERL w/ Cascade R-CNN           & \textcolor{red}{0.988}   \\ 		\hline 	
\end{tabular}    
\end{adjustbox} 
\end{table}

\subsection{Datasets}
We conduct experiments on three challenging underwater datasets UTDAC2020 \cite{utdac}, Brackish dataset \cite{brackish}, and TrashCan \cite{hong2020trashcan} to validate the performance of our method.

\textbf{UTDAC2020} is one of the series dataset from Underwater Target Detection Algorithm Competition 2020. There are $5,168$ training images and $ 1,293$ validation images in UTDAC2020 dataset. It contains 4 classes: echinus, holothurian, starfish, and scallop. The images in the dataset contain 4 resolutions: $3840 \times 2160, 1920 \times 1080, 720 \times 405$, and $586 \times 480$. 

\textbf{Brackish} is the first annotated underwater image dataset captured in temperate brackish waters. It contains 6 classes: bigfish, crab, jellyfish, shrimp, small fish, and starfish. This dataset is manually annotated with a bounding box annotation tool, resulting in a total of 14518 frames with 25613 annotations. The training set, validation set, and test set are randomly split into 9967, 1467, and 1468 images, respectively. The image size is $960 \times 540$.

\textbf{TrashCan} comes from the J-EDI deepsea image electronic library managed by the Japan Agency of Marine Earth Science and Technology (JAMSTEC), which contains 1000 video data of different lengths from various sea areas in Japan. It is labeled with two versions, including TrashCan-Instance and TrashCan-Material, we use TrashCan-Material in the experiments. The TrashCan-Material contains 16 object categories such as trash, rov, bio, unknown, metal,
and plastic. There are $6, 008$ training images and $1,204$ validation images in the TrashCan dataset. The image size is $480 \times 270$.

\subsection{Experimental Details and Evaluation Metrics}
\textbf{Experimental Details.} Our method is implemented on MMdetection \cite{mmdetection}. For our main results, we use multi-scale training with the long edge set to $1300$ and the short edge set to $800$. The models are trained for 16 epochs with an initial learning rate of 0.005, and decrease it by 0.1 after 8 and 11 epochs, respectively. Our method is trained on a single NVIDIA GeForce RTX 2080Ti GPU and adopts the SGD for training optimization, where the weight decay is 0.0001 and the momentum is 0.9. In the experiments, no data augmentation except the traditional horizontal flipping is utilized. All other hyper-parameters follow the settings in MMdetection.

\textbf{Evaluation Metrics.} The main reported results in this paper follow standard COCO-style Average Precision (AP) metrics that include $AP_{50}$ (IoU = 0.5), $AP_{75}$ (IoU = 0.75) and AP. AP is measured by averaging over multiple IoU thresholds, ranging from 0.5 to 0.95 with an interval of 0.05. Besides, we also use $AP_S$, $AP_M$, and $AP_L$ as metrics, which correspond to the results on small, medium, and large scales, respectively.  

In the next section, we show the effectiveness of ERL-Net by applying it with RetinaNet, Faster R-CNN, and Cascade R-CNN on two underwater datasets and a generic dataset.

\begin{table}[!t]   
	\centering   
	\caption{Performance comparison on TrashCan dataset. The results with red and blue colors indicate the best and second-best results of each column, respectively. }   
	\label{tab3}   
	\begin{adjustbox}{max width=0.48\textwidth}    
	  \begin{tabular}{c|ccc}    \hline   
		Method &YOLOv4 \cite{bochkovskiy2020yolov4} &RetinaNet w/ FPN \cite{retinanet}& SABL w/ Cascade R-CNN \cite{sabl} \\ 
		$AP_{50}$ & 0.559 & 0.362 &0.543  \  \\    \hline   
	  Method & Boosting R-CNN \cite{song2023boosting} & YOLOTrashCan\cite{zhou2022yolotrashcan} & Ours \\   
	  $AP_{50}$ & 0.560 &\textcolor{blue}{0.586}& \textcolor{red}{0.589} \\   
		\hline   
	  \end{tabular} 
	\end{adjustbox} 
  \end{table}

\subsection{Experiments on UTDAC2020 Dataset}

\begin{figure}[t] 	
	\centering 	
	\includegraphics[width=0.48\textwidth, keepaspectratio]{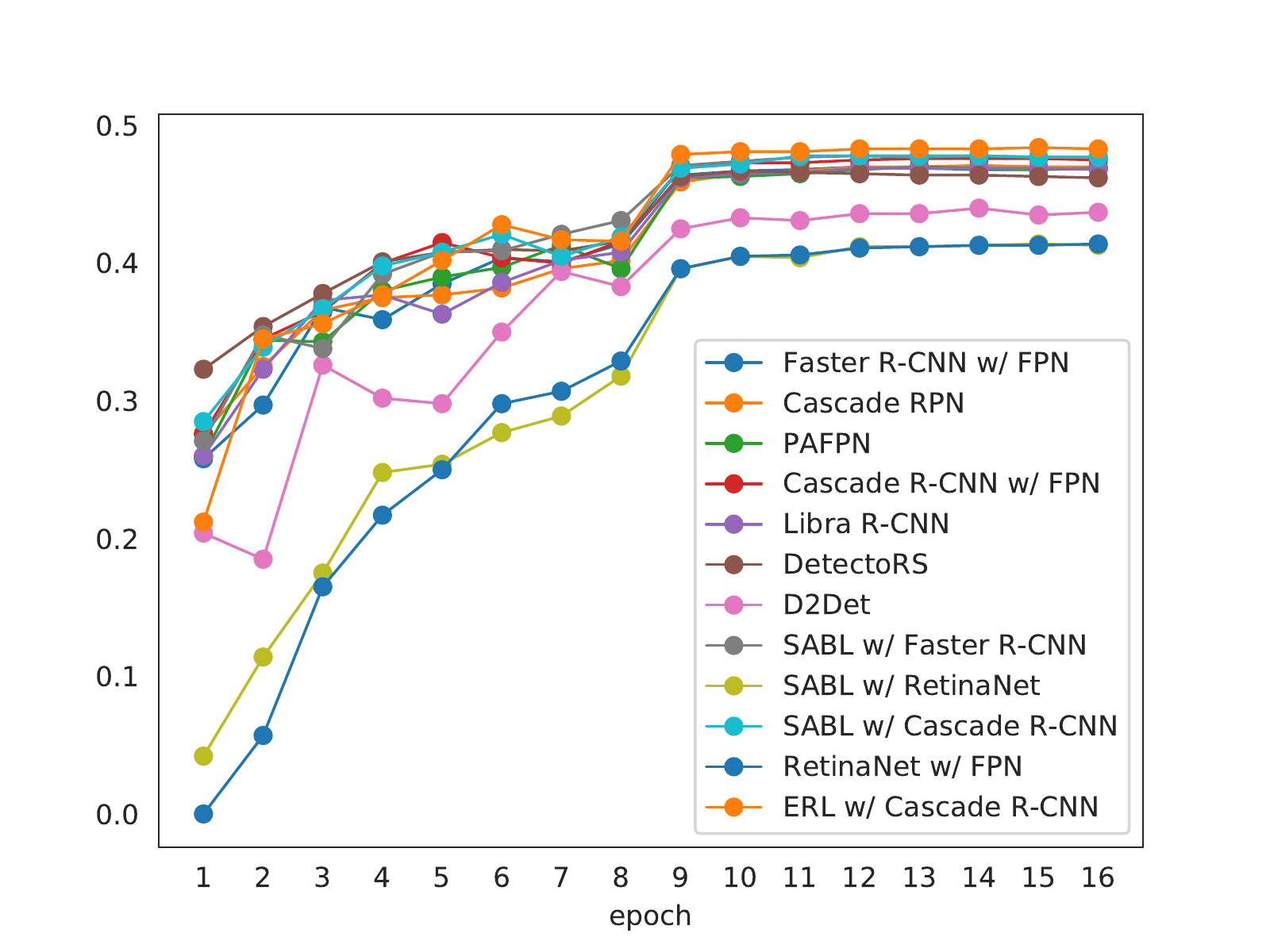} 	  \\ 	
	\caption{Comparison results of mAP between ERL-Net and other methods against epochs when ResNet-50 and FPN is used.} 	
	\label{fig:AP} 
\end{figure}

As shown in Table \ref{tab1} and Figure \ref{fig:AP}, ERL-Net improves the AP of RetinaNet by $3.7$ points (0.451 vs. 0.414) and Faster R-CNN by $1.4$ points (0.483 vs. 0.469). And we further apply ERL-Net to powerful Cascade R-CNN, ERL-Net improves the performance by $0.8$ points on this strong baseline (0.484 vs. 0.476). In addition, we further compare ERL-Net with other state-of-the-art two-stage object detection methods, such as Cascade RPN \cite{cascaderpn}, Libra R-CNN \cite{LibraRCNN}, PA-FPN \cite{pafpn}, DetectoRS \cite{detectors}, and Side-Aware Boundary localization (SABL) \cite{sabl}. Our ERL-Net significantly outperforms existing approaches by achieving an AP of $0.484$ with Cascade R-CNN. Further, compared with the latest SOTA method SABL and DetectoRS, the performance of ERL-Net achieves $0.6$ points higher than SABL and $2.0$ points than DetectoRS. The results show that ERL-Net has a better ability of precise location and classification than other methods in underwater object detection. 

\begin{figure}[t] 	 	
	\centering 	 	
	\includegraphics[width=0.48\textwidth, keepaspectratio]{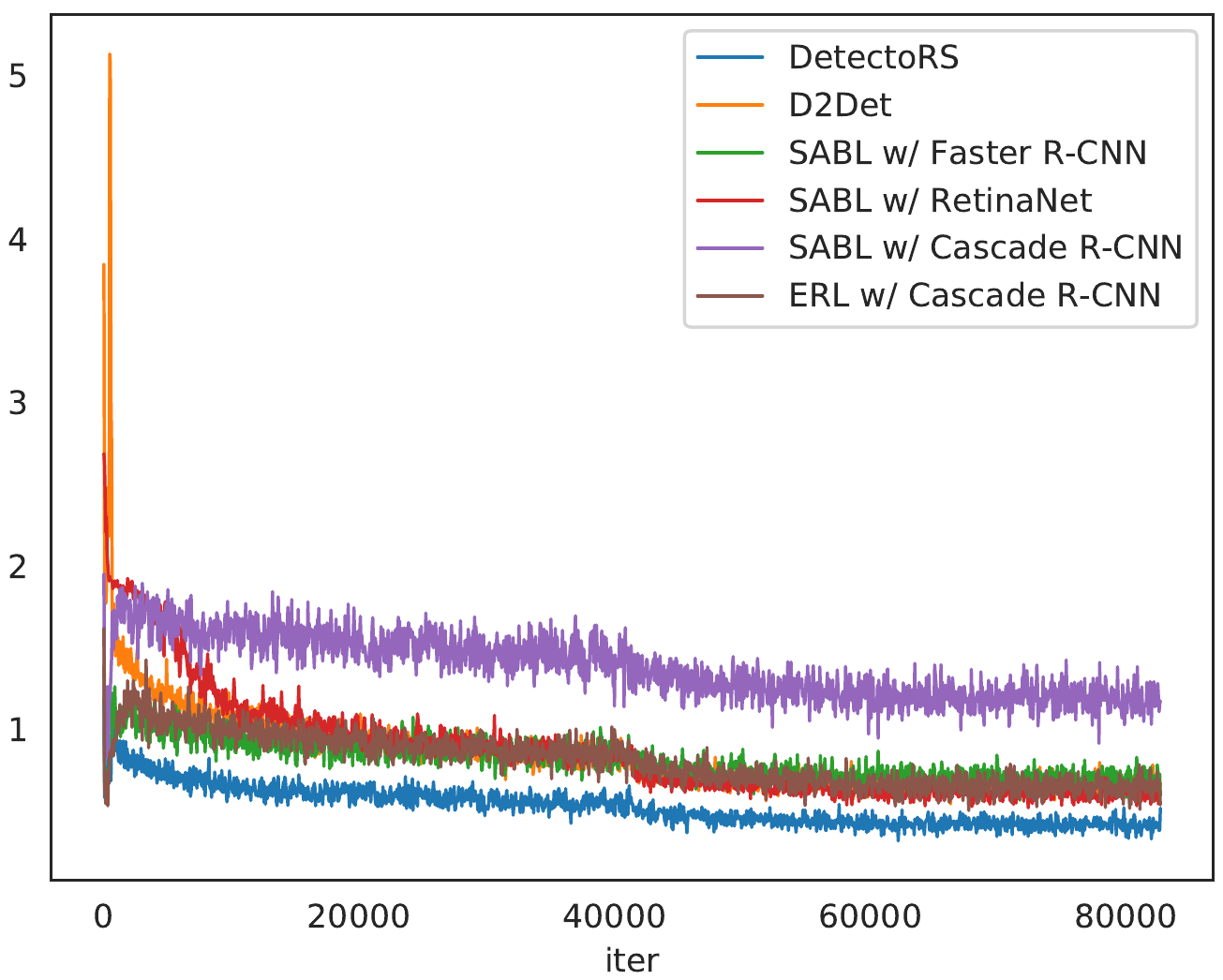} 	  \\ 	 	
	\caption{Comparison results of training loss between ERL-Net and other methods against epochs when ResNet-50 and FPN is used.} 	 	
	\label{fig:loss}  
\end{figure}

Apart from two-stage object detection methods, we also compare with single-stage detectors, such as SSD\cite{SSD}, FSAF \cite{fsaf}, and NAS-FCOS \cite{nasfcos}. Without bells and whistles, ERL-Net w/ RetinaNet achieves $0.451$ AP and brings 3.7 points higher AP than FSAF. And the AP of ERL-Net is only 0.7 points lower than NAS-FCOS. In addition, Figure \ref{fig:loss} shows the convergence speed of training, our method is quite simple to incorporate into two-stage or one-stage methods to achieve superior performance improvement without the sacrifice of too much inference time.

\subsection{Experiments on Brackish Dataset}
Next, we comprehensively evaluate ERL-Net on the Brackish Dataset. Table \ref{tab2} shows that the accuracy of the baseline algorithm \cite{brackish} on the Brackish dataset can only reach 0.851 AP. ERL-Net achieves a clear performance improvement. As shown in Table \ref{tab2}, ERL-Net with RetinaNet, Faster R-CNN, and Cascade R-CNN improve the baseline model by $14.2, 15.10$, and $15.10$, respectively. And we also compare with other SOTA methods, ERL-Net achieves 0.2 points higher than SABL w/ Faster R-CNN, 0.1 points higher than SABL w/ Cascade R-CNN, and 0.7 points higher than D2Det. The experimental results reveal the advantages of our method among advanced localization and classification pipelines.

\subsection{Experiments on TrashCan Dataset}
The comparison results on TrashCan are shown in Table \ref{tab3}. ERL-Net achieves 58.90\% mAP and surpasses the recently SOTA method YOLOTrashCan \cite{zhou2022yolotrashcan}. And we also compare with other methods, ERL-Net achieves 0.046 points higher than SABL w/ Faster R-CNN, 0.227 points higher than RetinaNet, and 0.029 points higher than Boosting R-CNN. The results indicate that our method is capable of adapting to different bodies of water, and even in cases where there are significant changes in the water environment, our method can still achieve good performance.

\subsection{Ablation Study}
In this section, we validate the effects of different components in the proposed method on the UTDAC2020 dataset.

\subsubsection{Analysis of Multi-level Feature Aggregation Module}
To explore the contribution of the multi-level feature aggregation module, we derive two settings: with/without multi-level feature fusion strategy. The results (+FA) in Table \ref{tab4} clearly show that the multi-level feature fusion is necessary for boosting performance. And we also conduct several experiments to explore different aggregation strategies. As shown in Table \ref{tab5}, ``Low 3 levels'' means that only aggregating the low-level features (conv1-3), ``High 3 levels'' means that only aggregating the high-level features (conv3-5). The results show that integrating only low-level features or only high-level features performs worse than our multi-level feature aggregation module. Since low-level features contain abundant details for refining the location of objects, but at the same time introduce a lot of background distraction.

\subsubsection{Impact of Wide and Asymmetric Receptive Field Module}
To investigate the importance of the WA-RFB module, we add it on the basis of a multi-level feature aggregation module. As shown in Table \ref{tab4}, adding the WA-RFB module is better than not adding it. This is because WA-RFB module can expand the receptive field and focus on multi-scale global contextual information, which is important for underwater object detection.

In addition, we also compare WA-RFB with the original RFB \cite{RFB}, as shown in Table \ref{tab4}. The mAP of adding WA-RFB is $0.478$, while mAP of adding RFB is $0.475$. There is no change in the mAP with and without RFB. The WA-RFB surpasses RFB on the all metrics. It means that the WA-RFB is beneficial for locating underwater objects. 

\begin{table*}
	\centering
	\caption{Ablation study of the proposed ERL-Net on the validation set of UTDAC2020, the baseline represents the result of Faster R-CNN with FPN. ``FA'' represents the feature aggregation module. ``WA-RFB'' denotes the wide and asymmetric receptive field block. ``CA'' and ``EGA'' represent the channel attention and edge-guided attention.}
	\label{tab4}
	\begin{adjustbox}{max width=\textwidth}
	\begin{tabular}{l|lllll|lll|llll} 
		\hline
		Methods  & FA            & RFB        & WA-RFB & EGA            & CA            & mAP             & $AP_{50}$       & $AP_{75}$       & holo.     & ech.         & sca.         & star.         \\ 
		\hline
		Baseline \cite{brackish} &               &               &     &               &               & 0.469           & 0.820           & 0.491           & 0.385           & 0.485           & 0.481           & 0.526            \\
		+FA      & \cmark  &               &     &               &               & 0.475           & 0.824           & 0.500           & 0.404           & 0.482           & 0.481           & 0.532            \\
		+RFB     &    \cmark           &     \cmark          &   &               &               &  0.475   & 0.817         &   0.507 &      0.394           &  0.494         &  0.479    &   0.534           \\
		+WA-RFB  & \cmark  &   &   \cmark  & &  & 0.478           & 0.828           & 0.507 & \textbf{0.405}           & 0.485           & \textbf{0.490}           & 0.531            \\
		+only EGA      & \cmark  &  &  \cmark    & \cmark  & & 0.481           & 0.830           & \textbf{0.514}  & 0.402           & \textbf{0.501}  & 0.482           & 0.538            \\
		+only CA      & \cmark  &   &  \cmark   &  & \cmark  & 0.477  & 0.825 & 0.502 & 0.398    & 0.493  & 0.483           & 0.533   \\
		+EGA\&CA      & \cmark  &   &  \cmark   & \cmark  & \cmark  & \textbf{0.483}  & \textbf{0.832}  & 0.513           & \textbf{0.405}  & 0.498           & 0.488  & \textbf{0.541}   \\
		\hline
		\end{tabular}
	\end{adjustbox}
\end{table*}
\begin{table} 	
	\centering 	
	\caption{Comparison of different feature aggregation strategies on the UTDAC2020 dataset.} 	
	\label{tab5} 	
	\begin{tabular}{llll}  		
		\hline 		Settings      & AP & $AP_{50}$ & $AP_{75}$ \\  		\hline 		Low 3 levels  &  0.467  &   0.811    &    0.478    \\ 		High 3 levels &  0.471  &   0.813    &    0.480    \\ 		FA (ours)     & \textbf{0.475}   &  \textbf{0.824}  &  \textbf{0.500} \\ 		\hline 	
	\end{tabular} 
\end{table}

\begin{figure*}[t]
	\centering
	\includegraphics[width=0.95\textwidth, keepaspectratio]{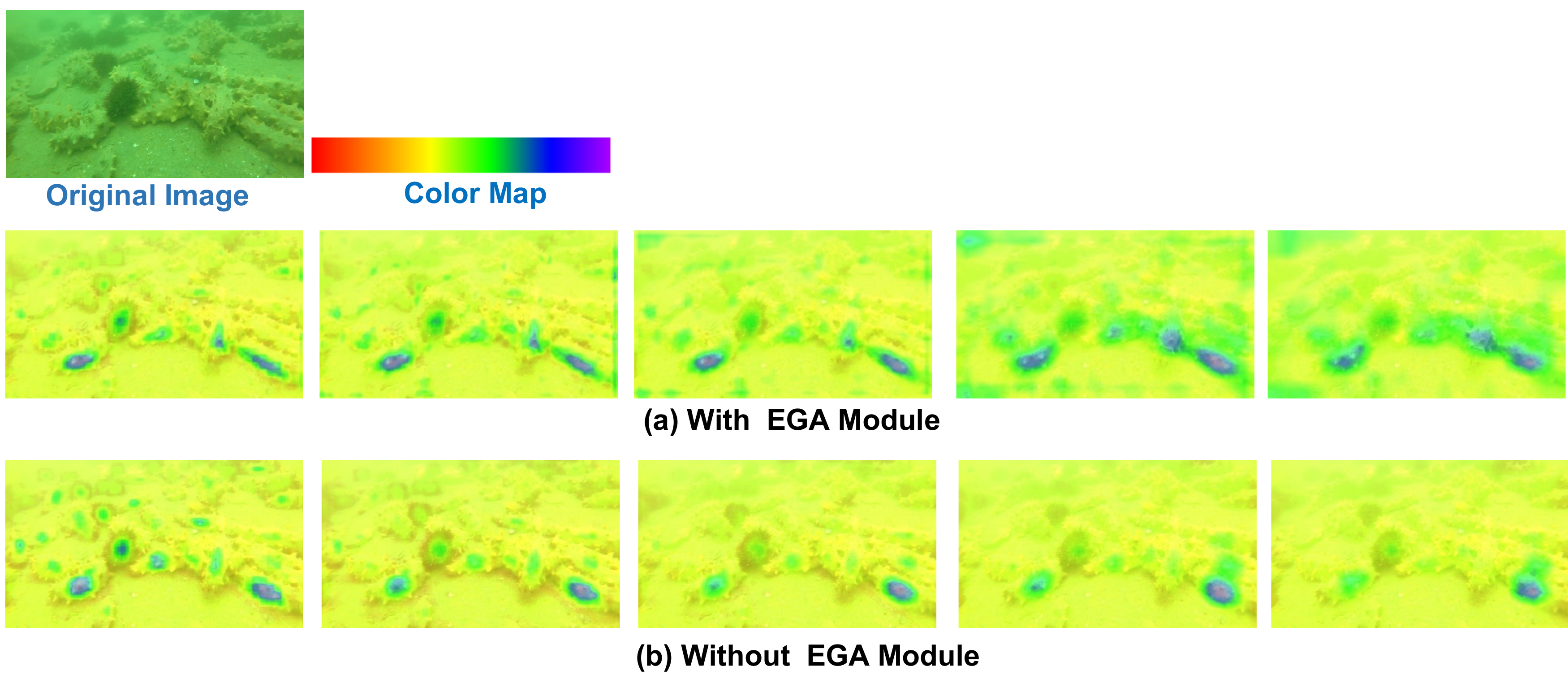} . 
	\caption{Visualization comparison results of feature map with (second row) or without (third row) EGA module. The first to fifth columns show the feature map of each stage of FPN. The colormap represents the attention degree of the feature (the higher the better). The response size of the feature map is mapped to the original image, allowing us to more intuitively understand the effect of the EGA. As we can observe, EGA enforces the model to focus on the shape of underwater objects. The feature map with EGA demonstrates the prominence of the edge shape.}
	\label{fig:heatmap}
\end{figure*}

\begin{figure*}[t]
	\centering
	\includegraphics[width=0.92\textwidth,height=\textheight,keepaspectratio]{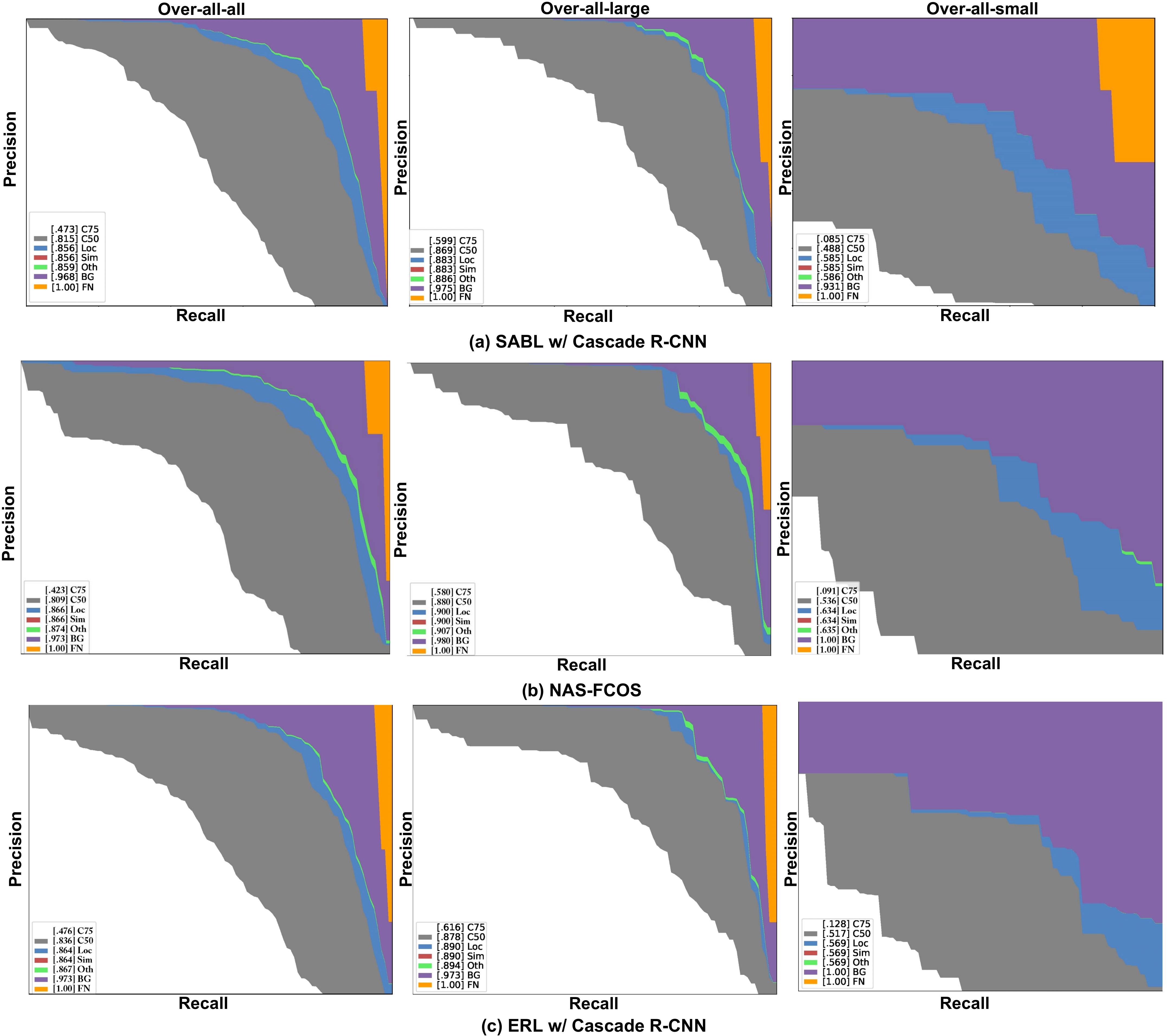} 
	\caption{Error analysis plots showing a comparison of our ERL-Net (bottom row) with SABL (top row) and NAS-FCOS (middle row) across three categories, on the overall (first column), the large-sized objects (second column), and small-sized objects (the last column). As defined in \cite{COCO}, a series of precision-recall curves with different evaluation settings is shown in each sub-image plot.}
	\label{fig:error}
\end{figure*}
\begin{figure*}[t]
	\centering
	\includegraphics[width=0.9\textwidth, height=\textheight, keepaspectratio]{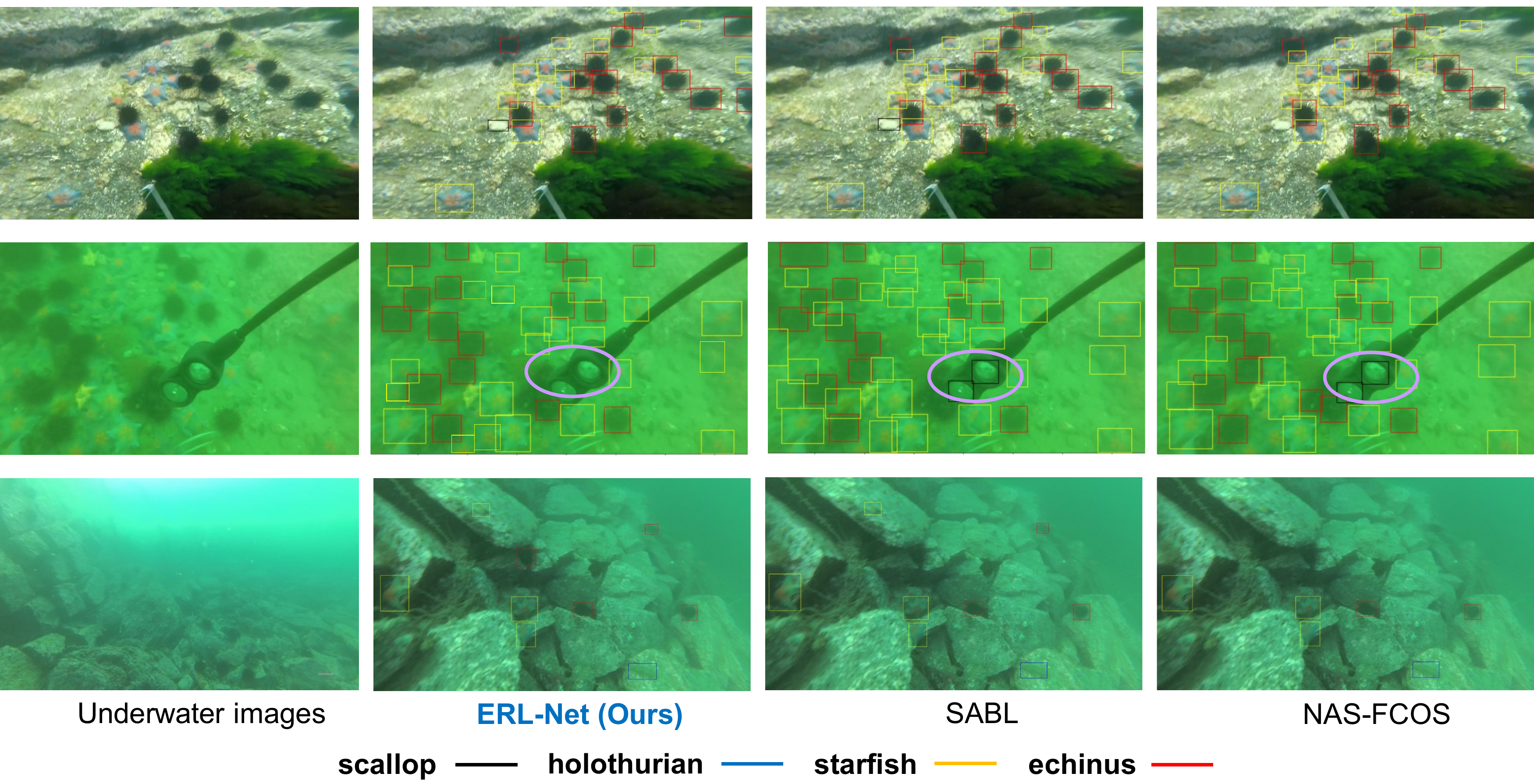}  
	\caption{Qualitative comparison results on the UTDAC2020 dataset. Each color of a box belongs to an object class. Our method ERL-Net w/ Cascade R-CNN can locate the small underwater objects accurately in various complex underwater scenarios, including small dense objects (top row), low-contrast scenarios (middle row), and complex backgrounds (the last row). The pink circles indicate obvious examples of errors. Other methods even mistake diving equipment for echinus.}
	\label{fig:UTDAC}
\end{figure*}
\begin{figure*}[t]
	\centering
	\includegraphics[width=0.9\textwidth, keepaspectratio]{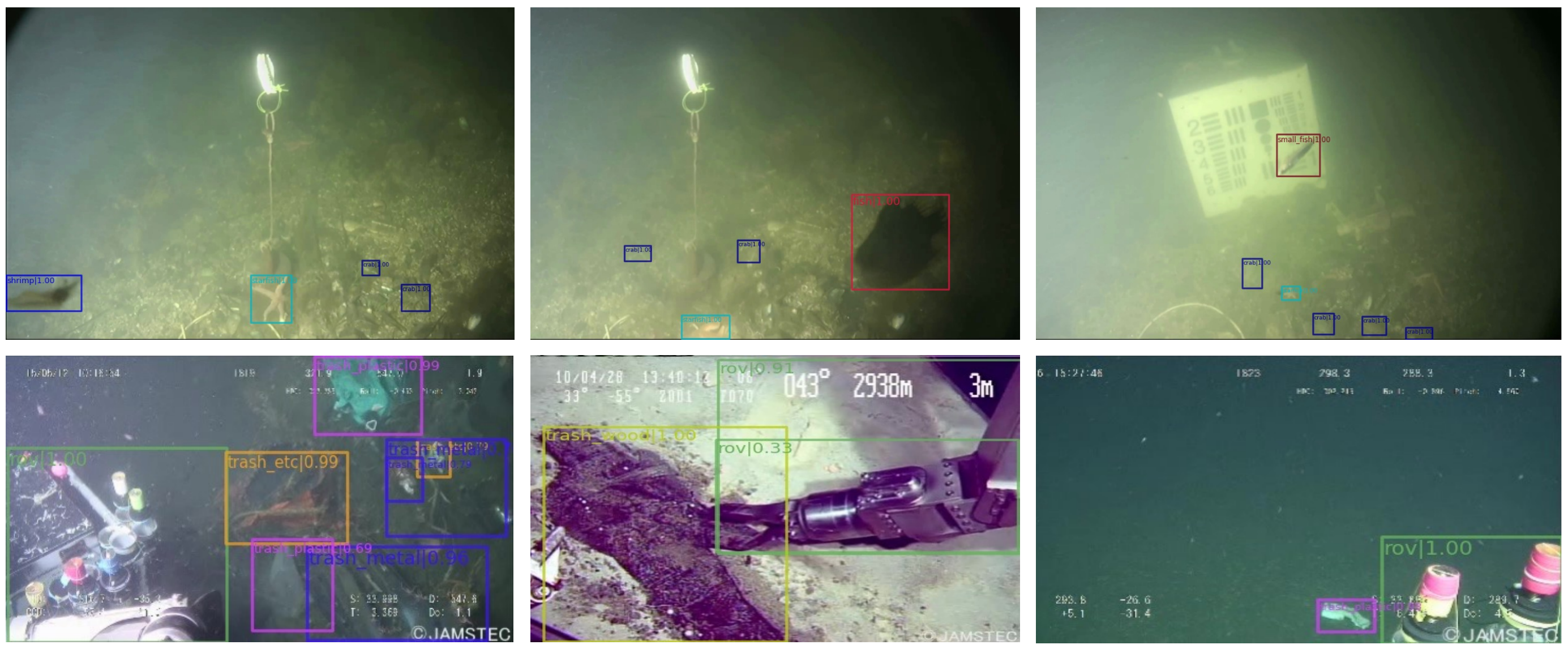}. 
	\caption{Qualitative detection results of ERL-Net on the Brackish \cite{brackish} (the first row) and TrashCan \cite{hong2020trashcan} (the second row). Best viewed in color and with zoom.}
	\label{fig:brackish}
\end{figure*}

\subsubsection{Impact of Edge-guided Attention Module}
In this subsection, we explore the role of edge information, which is also our main idea. As shown in Table \ref{tab4}, adding the edge information (+EGA) performs better than other settings. These improvements demonstrate that the edge information is very useful for the underwater object detection task. Compared with the baseline model without EGA module, after adding the edge attention, edge information can help refine the boundaries of underwater objects. 

In order to further illustrate that our method does learn edge information, we also visualize the feature map of every stage of FPN (described in Sec.3.4.) with or without adding EGA module. These attention maps are calculated based on the activation degree of each spatial unit in the convolution layer, which basically reflects where the network focuses most of its attention in order to classify and locate the input image. The results are shown in Figure \ref{fig:heatmap}, from which we can see that after adding edge information, the proposed method can depict more delicate edges and focus on the whole shape of the target, not just the center point.

We also investigate the importance of the adaptive channel-wise attention mechanism. When only adding the CA (without EGA), the mAP is $0.477$. However, while both use CA and EGA, the mAP is $0.484$. The results show that both edge-guided attention and channel-wise attention are meaningful for underwater object detection with the assistance of edge cues. The guidance of edge information is helpful in pay attention to the effect of edge on recognition, and the channel attention mechanism suppresses the interference of some irrelevant information.

\subsection{Error Analysis and Qualitative Results}
\textbf{Error Analysis:} To further analyze the proposed method, we first utilize the error analysis tool provided by COCO \cite{COCO}. Figure \ref{fig:error} shows the error plots on the UTDAC2020 dataset for ERL-Net combined with Cascade R-CNN. The plots in each sub-image represent a series of precision-recall curves with various evaluation settings. The curves in the plots are as follows: (1) \textbf{C75}: PR at IoU=0.75 (AP at strict IoU), area under curve corresponds to $AP_{75}$ metric; (2) \textbf{C50}: PR at IoU=0.50 (AP at PASCAL IoU), area under curve corresponds to $AP_{50}$ metric; (3) \textbf{Loc}: PR at IoU=0.10 (localization errors ignored, but not duplicate detections). All remaining settings use IoU=0.10; (4) \textbf{Sim}: PR after super-category false positives (fps) are removed; (5) \textbf{Oth}: PR after all class confusions are removed; (6) \textbf{BG}: PR after all background (and class confusion) fps are removed; (7) \textbf{FN}: PR after all remaining errors are removed (trivially AP=1.0). The area under each curve is shown in brackets in the legend.
In the case of overall results (on the left), our ERL-Net w/ Cascade R-CNN achieves 0.476 AP at the strict $AP_{75}$, while SABL w/ Cascade R-CNN obtains 0.485 AP and NAS-FCOS obtains 0.423 AP. Though our method is slightly lower than SABL at strict $AP_{75}$, our method achieve a prominent improvement at AP@[0.5:0.05:.95] (\textbf{0.484} vs. 0.478, shown in Table \ref{tab1}) and $AP_{50}$ (\textbf{0.836} vs. 0.815). In the case of overall-all-small, our method provides a gain of near 2.5\%  and 3.7 \% by achieving 0.128 AP at the strict metric of $AP_{75}$, compared to 0.085 by SABL and 0.091 by NAS-FCOS. Our method shows the superior performance on the small objects.

\textbf{Qualitative Results:} We present predicted examples of our method and several SOTA algorithms. Figure \ref{fig:UTDAC} shows the comparison results on the UTDAC2020 dataset. Visualizations cover various challenging scenarios, including small dense objects, low-contrast scenes, and complex backgrounds. As can be seen, ERL-Net performs better on dense small object localization and classification. It is worth mentioning that thanks to the discriminative edge features, our result could avoid some false detection problems. For instance, for the second example, due to the underwater detector device's influence, other SOTA methods even mistake the underwater detector equipment for echinus. However, benefiting from the complementary edge features, our method performs better and eliminates the false detection. We also present the qualitative results of ERL-Net on the Brackish and TrashCan dataset, shown in Figure \ref{fig:brackish}. ERL-Net is outstanding in the detection of small detection in the low-contrast environment (the first row) and densely arranged objects (the second row). 

\section{Conclusion}
In this work, we propose an Edge-guided Representation Learning Network (ERL-Net) for underwater object detection. Underwater object detection is challenging due to low-contrast, small objects, and mimicry of aquatic organisms. To address these challenges, we propose a novel edge-guided attention module to extract information cues from the edge of objects. Then, we integrate the local edge information and global low-level content information using a feature aggregation module. Furthermore, we introduce a wide and asymmetric receptive field block that enlarges the feature receptive filed and enhances the multi-scale feature discriminability. We combined the proposed ERL-Net with various popular object detection pipelines, including the one-stage method RetinaNet and two-stage methods Faster R-CNN and Cascade R-CNN. Experiments on three challenging underwater datasets demonstrate that ERL-Net outperforms other state-of-the-art methods on underwater object detection. 

\bibliographystyle{elsarticle-num} 
\bibliography{ref}
\end{document}